%% file: main.tex
\documentclass[10pt,twocolumn,letterpaper]{article}

\usepackage{cvpr}              


\usepackage[accsupp]{axessibility}
\usepackage{algorithm}
\usepackage{algpseudocode}
\usepackage{multirow}
\usepackage{colortbl}
\usepackage{adjustbox}
\usepackage{tabularx, booktabs}
\usepackage{caption}
\usepackage{subcaption}
\usepackage{tcolorbox}
\usepackage{enumitem}
\usepackage{pgfplots}
\pgfplotsset{compat=1.18}
\usepgfplotslibrary{fillbetween}
\usepackage{tikz}

\usepackage{pifont}
\newcommand{\cmark}{\ding{51}}
\newcommand{\xmark}{\ding{55}}

\newcommand{\sftypef}[1]{{\textsf{\footnotesize #1}}}

\usepackage{amsmath} 
\renewcommand{\eqref}[1]{\textup{(\ref{#1})}}
\usepackage{color, colortbl}
\usepackage{float}
\usepackage{bm}
\usepackage{cuted}
\usepackage{capt-of}

\usepackage{amsthm}
\theoremstyle{plain}
\newtheorem{theorem}{Theorem}[section]

\newtheorem{lemma}[theorem]{Lemma}

\newtheorem{assumption}[theorem]{Assumption}

\definecolor{cBlue}{HTML}{1852CC}
\definecolor{cBlue2}{HTML}{3fb8b8}
\definecolor{cBlue3}{HTML}{02ffff}
\definecolor{cBlue4}{HTML}{00b5ff}
\definecolor{cRed}{HTML}{D62728}
\definecolor{cRed2}{HTML}{ED521F}
\definecolor{cRed3}{HTML}{F69C40}
\definecolor{cRed4}{HTML}{fa4d3a}
\definecolor{cGreen}{HTML}{2CA02C}
\definecolor{cGreen2}{HTML}{3fdf3f}
\definecolor{cPink}{HTML}{ED1FD2}
\definecolor{cWhite}{HTML}{ffffff}
\definecolor{Violet}{HTML}{b05cff}
\definecolor{Gray}{gray}{0.9}
\definecolor{Salmon}{HTML}{FF7E79}
\definecolor{Orchid}{HTML}{7A81FF}
\definecolor{cBlue5}{HTML}{409BA0}
\definecolor{cPink2}{HTML}{CB2CED}
\definecolor{cPink3}{HTML}{ED1D81}
\definecolor{PastelPink}{HTML}{FC94AF}

\definecolor{cvprblue}{rgb}{0.21,0.49,0.74}
\usepackage[pagebackref,breaklinks,colorlinks,allcolors=cvprblue]{hyperref}

\def\paperID{16006} 
\def\confName{CVPR}
\def\confYear{2026}

\title{ChimeraLoRA: Multi-Head LoRA-Guided Synthetic Datasets}

\author{
Hoyoung Kim\textsuperscript{1} \quad
Minwoo Jang\textsuperscript{1} \quad
Jabin Koo\textsuperscript{2} \quad
Sangdoo Yun\textsuperscript{3} \quad
Jungseul Ok\textsuperscript{1,2}\\ \\
$\text{Graduate School of AI, POSTECH}^1$, \quad 
$\text{Dept. of CSE, POSTECH}^2$, \quad
$\text{NAVER AI Lab}^3$\\
{\tt\small \url{https://cskhy16.github.io/chimeralora}}
}

\begin{document}
\maketitle

\input{Sections/0_abstract}

\input{Sections/1_introduction}

\input{Sections/2_related_work}
\input{Sections/3_preliminary}

\input{Sections/4_method}

\input{Sections/5_experiment}
\input{Sections/6_conclusion}
{
    \small
    \bibliographystyle{ieeenat_fullname}
    \bibliography{main}
}
\input{Sections/X_suppl}


\end{document}

%% file: Sections/0_abstract.tex
\begin{abstract}
Beyond general recognition tasks, specialized domains and fine-grained settings often encounter data scarcity, especially for tail classes. To obtain less biased and more reliable models under such scarcity, practitioners leverage diffusion models to supplement underrepresented regions of real data. Specifically, recent studies fine-tune pretrained diffusion models with LoRA on few-shot real sets to synthesize additional images. While an image-wise LoRA trained on a single image captures fine-grained details yet offers limited diversity, a class-wise LoRA trained over all shots produces diverse images as it encodes class priors yet tends to overlook fine details. To combine both benefits, we separate the adapter into a class-shared LoRA~$A$ for class priors and per-image LoRAs~$\mathcal{B}$ for image-specific characteristics. To expose coherent class semantics in the shared LoRA~$A$, we propose a semantic boosting by preserving class bounding boxes during training. For generation, we compose $A$ with a mixture of $\mathcal{B}$ using coefficients drawn from a Dirichlet distribution. Across diverse datasets, our synthesized images are both diverse and detail-rich while closely aligning with the few-shot real distribution, yielding robust gains in downstream classification accuracy.
\end{abstract}

%% file: Sections/1_introduction.tex
\section{Introduction}
While general recognition tasks enjoy abundant and class-balanced data, specialized domains often face data scarcity~\citep{tian2020rethink,Lee_2022_CVPR,Chowdhury_2021_ICCV} and long-tailed class distributions~\citep{liu2019large,cui2019class}.
For example, class rarity in fine-grained tasks can limit data collection, leaving only a few labeled images per class~\citep{10.1007/978-3-031-73247-8_12,guo2025focus}. 
Training under such data scarcity often causes models to overfit and learn decision boundaries biased toward majority classes, degrading generalization performance~\citep{cao2019learning}. To supplement limited data, recent work leverages generative priors in pretrained text-to-image diffusion models~\citep{rombach2022high} to synthesize additional training images by conditioning class names with text prompts~\citep{hesynthetic}. However, without guidance from real images, synthetic data easily drifts from the target distribution and lower downstream accuracy~\citep{taori2020measuring,hesynthetic}.

To narrow the real-to-synthetic gap, recent work exploits few-shot real images~\citep{kim2024datadream,kim2025loft,da2023diversified,hesynthetic}. Specifically, a training-free image-wise baseline initializes the diffusion process from features of a single reference image to synthesize samples close to that reference~\citep{hesynthetic}.
Beyond this, an image-wise variant embeds the reference into a diffusion model by fine-tuning lightweight low-rank adapters (LoRAs)~\citep{hu2022lora}, enabling the model to capture fine-grained details~\citep{kim2025loft}.
However, these image-wise approaches make it difficult to generate diverse images, as they rely on a single image.
To leverage the remaining images of the same class, a class-wise LoRA is fine-tuned on all shots to encode class-level priors and promote diversity, yet the resulting samples often overlook instance-specific details~\citep{kim2024datadream}.
This trade-off stems from adapting the diffusion model with LoRA at a single granularity, either an image or a class, motivating a unified image- and class-level adaptation.


\begin{figure*}[t!]
\centering
\includegraphics[width=\textwidth]{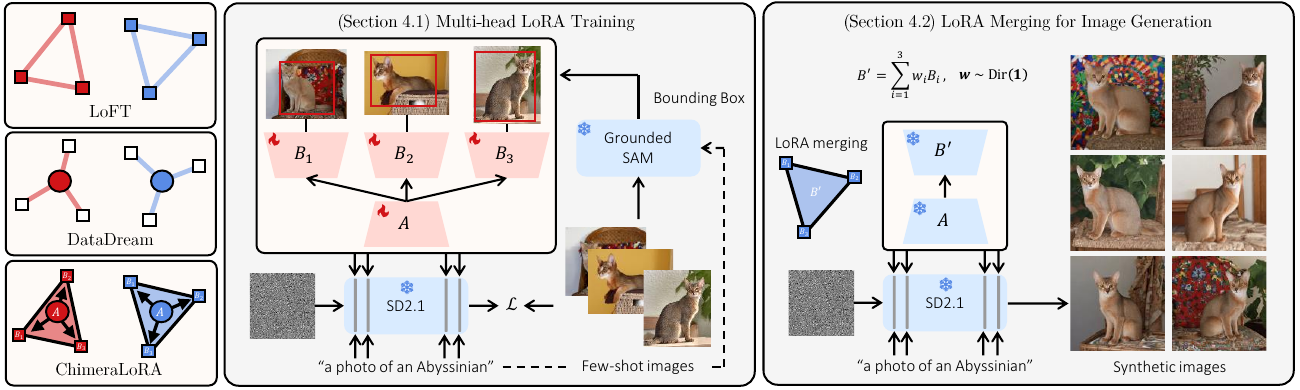}
\caption{
{\em An overview of the proposed method.}
(left) We synthesize images with a multi-head LoRA that integrates the strengths of image-wise LoRA (LoFT~\citep{kim2025loft}) and class-wise LoRA (DataDream~\citep{kim2024datadream}). The blue and red regions indicate where LoRA is applied during generation. (center) Given few-shot images, we fine-tune the multi-head LoRA while preserving bounding boxes obtained from Grounded-SAM~\citep{liu2023grounding,kirillov2023segment}. (right) We merge LoRA heads using weights sampled from a Dirichlet distribution to obtain diverse synthetic images.
}
\vspace{-1.5mm}
\label{fig:main}
\end{figure*}


To generate synthetic images that are both diverse and fine-grained, we adopt a multi-head LoRA architecture~\citep{tian2024hydralora,guo2025selective}. 
Specifically, LoRA~\citep{hu2022lora} approximates updates to a large weight matrix as the product of two low-rank matrices, LoRA~$A$ and LoRA~$B$.
We give these two LoRAs distinct roles by sharing a single LoRA~$A$ across all few-shot images and assigning each image its own LoRA~$B$. In this way, the shared LoRA~$A$ captures class-level priors that drive diverse generation, while the per-image LoRA heads~$\mathcal{B}$ encode instance-specific details.
For training, we freeze the base diffusion model and jointly fine-tune the LoRA~$A$ and $B$.
To promote coherent class semantics in the shared $A$, we propose a semantic boosting technique that utilizes bounding boxes localized by Grounded Segment Anything Model~\citep{liu2023grounding,kirillov2023segment}.

At generation time, we fix $A$ and mix multiple heads from $\mathcal{B}$ with nonnegative coefficients sampled from a Dirichlet distribution.
Thanks to the shared $A$ and a different Dirichlet-weighted head for each generated image, the synthesized images exhibit both fine-grained details and diversity. 
In addition, our semantic boosting drives coverage of the full visible extent of a target class, rather than a partial visibility.
Across diverse classification tasks, including realistic medical domains and long-tailed scenarios, our synthetic images are not only distribution-aligned to the few-shot references, yielding robust gains over baselines, but also qualitatively diverse and detailed.
Figure~\ref{fig:main} compares our method with prior work and outlines the overall pipeline.


Our main contributions are summarized as follows:
\begin{itemize}
\vspace{1mm}
\item We present a multi-head LoRA framework in which LoRA~$A$ encodes class-level priors and LoRA heads~$\mathcal{B}$ capture instance-specific details, producing diverse and fine-grained synthetic images (Section~\ref{sec:alt}).
\vspace{1mm}
\item Our synthetic datasets generally improve downstream accuracy across various benchmarks, including specialized domains and long-tailed settings (Section~\ref{exp:few-shot}).
\vspace{1mm}
\item We generate synthetic images aligned with the real few-shot distribution and analyze the synthetic-to-real gap both qualitatively and quantitatively (Section~\ref{sec:analyses}).
\end{itemize}

%% file: Sections/2_related_work.tex
\section{Related Work}

\paragraph{Few-shot guided synthetic datasets.}
Recent methods for synthetic dataset generation leverage real images as guidance rather than relying solely on text prompts. For example, IsSynth~\citep{he2023is} conditions on the latent space of real images to improve performance, and DISEF~\citep{da2023diversified} perturbs real-image latents for generation.
Recently, LoFT~\citep{kim2025loft} trains image-wise LoRA adapters and mixes their contributions at sampling. While these single-image approaches capture fine-grained details, they overlook the broader class distribution. To improve class-level coverage, DataDream~\citep{kim2024datadream} trains class-wise LoRA adapters on all shots of a class, emphasizing generality over image-level fidelity. To combine both strengths, we introduce a shared LoRA~$A$ for class-level priors and LoRA heads~$\mathcal{B}$ for instance-specific details, unifying class-level generality with image-level fidelity.

\paragraph{Multi-head LoRA architectures.}
Even in a single-head LoRA~\citep{hu2022lora}, recent analysis identifies an asymmetric role: LoRA~$A$ acts as a simple projection agnostic to the input distribution, whereas LoRA~$B$ tends to capture the input data distribution~\citep{pmlr-v235-zhu24c}. Building on this observation, multi-head designs explicitly allocate task or instance specific capacity to multiple LoRA~$B$ heads while sharing LoRA~$A$. Specifically, HydraLoRA~\citep{tian2024hydralora} employs expert routing over LoRA~$B$ heads, FedSA-LoRA~\citep{guo2025selective} keeps local LoRA~$B$ per client to address heterogeneity in federated learning, and AsymLoRA~\citep{wei2025asymlora} applies this paradigm to multimodal instruction tuning. Following these developments, we adopt an asymmetric multi-head LoRA architecture to fine tune diffusion models on few shot images.

\paragraph{Semantic preservation in data augmentation.}
To address the challenge of semantic preservation during data augmentation, several methods propose alternatives beyond standard transformations. For instance, KeepAugment~\citep{gong2021keepaugment} utilizes saliency maps to preserve informative regions, while ObjectCrop~\citep{mishra2022objectaware} and ContrastiveCrop~\citep{peng2022crafting} aim to obtain reliable positive samples in contrastive learning by leveraging object proposals and semantic-aware localization, respectively. With the emergence of Segment Anything Model (SAM)~\citep{kirillov2023segment}, SAMAug~\citep{zhang2023input} simply combines raw images with SAM-generated masks for medical image segmentation. However, these approaches do not explicitly ensure object integrity under cropping. In contrast, we focus on generating images with complete objects for synthetic datasets by leveraging bounding boxes from Grounded-SAM~\citep{liu2023grounding,kirillov2023segment}.

%% file: Sections/3_preliminary.tex
\begin{figure*}[t!]
\centering
\includegraphics[width=\textwidth]{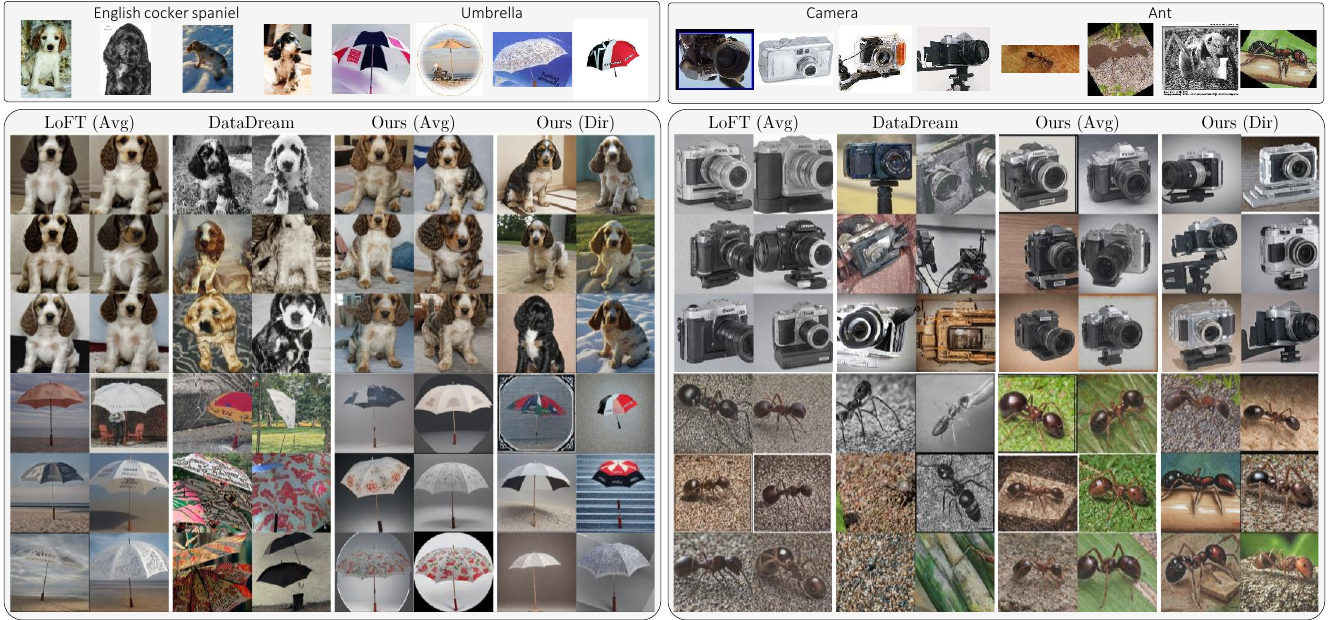}
\caption{
{\em Qualitative results of synthetic images.} (top) Four real images per class. (bottom) Synthetic images generated with LoRA based methods. For the camera class, LoFT (image-wise LoRA) shows low diversity with near duplicate single viewpoint shots, while DataDream (class-wise LoRA) increases diversity but lowers fidelity, often failing to render a camera. Our multi-head LoRA produces accurate cameras across varied viewpoints. Here, \sftypef{Avg} merges heads with uniform weights and \sftypef{Dir} uses Dirichlet sampled weights.
}
\label{fig:qual}
\vspace{-3mm}
\end{figure*}

\section{Preliminaries}
To supplement few shot examples, we leverage the generative power of large-scale diffusion models combined with parameter-efficient LoRA adaptation.
Therefore, as preliminary background, we first introduce the fundamentals of latent diffusion models in Section~\ref{sec:ldm}, and then review previous approaches that employ LoRA for synthetic image generation in Section~\ref{sec:lora}.

\subsection{Latent Diffusion Models}
\label{sec:ldm}
Latent diffusion models (LDMs), such as Stable Diffusion~\citep{rombach2022high}, are probabilistic generative models designed to generate high-resolution images conditioned on text prompts $y$. Let $\mathcal{D}$ be a dataset of image-text pairs. For $(x, y) \in \mathcal{D}$, let an encoder $\mathcal{F}$ map the image $x$ into a latent representation $z=\mathcal{F}(x)$. The forward diffusion process progressively adds Gaussian noise $\epsilon$ sampled from a standard normal distribution $\mathcal{N}(0,1)$ to the latent $z$ over $t$ steps, resulting in increasingly noisy latent variables $z_t$. The reverse process learns to iteratively remove this noise, conditioned on the text prompt $y$. Specifically, an intermediate representation $\tau(y)$ obtained from a pretrained text encoder $\tau$ is provided to the cross-attention layers of the UNet~\citep{ronneberger2015u} to guide the denoising process. To achieve this, the conditional LDM parameterized by $\theta$ is trained with the objective as follows:
\begin{align}
\min_{\theta} \mathbb{E}_{(x,y) \sim \mathcal{D}, \epsilon \sim \mathcal{N}(0, 1), t} \big[ \lVert \epsilon - \epsilon_{\theta}(z_{t}, t, \tau(y)) \rVert_{2}^{2} \big] \;,
\end{align}
where $t$ is uniformly sampled from $\{1,\dots,T\}$ with $T$ the number of diffusion timesteps. To generate a synthetic image, an initial latent noise $z_T$ is iteratively denoised conditioned on the text prompt $y$, and the resulting latent $z_0$ is decoded by a decoder $\mathcal{G}$ to produce the final image $x'=\mathcal{G}(z_0)$. While previous works~\citep{saharia2022photorealistic, ramesh2022hierarchical} primarily focus on the visual quality of individual generated images, we investigate whether the collection of synthetic images produced by LDMs can serve as effective training datasets for downstream tasks.

\subsection{Single-head LoRA-Guided Synthetic Datasets}
\label{sec:lora}
Generating synthetic datasets solely from a pretrained LDM conditioned on text class prompts $c \in \mathcal{C}$, where $\mathcal{C}$ denotes the set of target classes, often results in significant distribution shifts relative to the downstream target task~\citep{fan2024scaling, hesynthetic}. 
Recent work mitigates this issue by few-shot guidance, assuming access to a small labeled dataset $\mathcal{D}_\text{fs} = \{(x_i,y_i)\}_{i=1}^{K|\mathcal{C}|}$, containing $K$ examples per class~\citep{da2023diversified}.
In this setting, LoRA~\citep{hu2022lora} has been employed to efficiently adapt a pretrained LDM $\theta$ to the few-shot dataset $\mathcal{D}_{\text{fs}}$. Specifically, given a pretrained weight matrix $W_0 \in \mathbb{R}^{d_1 \times d_2}$, LoRA introduces two trainable low-rank matrices $B \in \mathbb{R}^{d_1 \times r}$ and $A \in \mathbb{R}^{r \times d_2}$, with rank $r \ll \min(d_1, d_2)$. Keeping the LDM parameters $\theta$ fixed, $A$ and $B$ are jointly optimized on a subset $\mathcal{D}' \subseteq \mathcal{D}_\text{fs}$ as follows:
\begin{align}
\!\! \min_{A,B} \mathbb{E}_{(x,y) \sim \mathcal{D}', \epsilon \sim \mathcal{N}(0, 1), t} \big[ \lVert \epsilon - \epsilon_{\theta, A, B}(z_{t}, t, \tau(y)) \rVert_{2}^{2} \big]\;.
\end{align}
Depending on the choice of the subset $\mathcal{D}'$, the resulting LoRA adapters capture visual variability at different granularities.
An image-wise LoRA is trained on a single-image subset $\mathcal{D}'=\{(x,y)\}$, capturing instance-specific features and often yielding high fidelity, yet offering limited coverage of the class distribution.
In contrast, a class-wise LoRA is optimized per class $c$ using $\mathcal{D}'=\{(x,y) \in \mathcal{D}_\text{fs} \mid y=c \}$, encoding class priors and promoting broader diversity, but it overlooks instance-level details.
Figure~\ref{fig:qual} shows that single granularity methods degrade quality: LoFT~\citep{kim2025loft} (image-wise) yields low diversity, whereas DataDream~\citep{kim2024datadream} (class-wise) shows low fidelity.

%% file: Sections/4_method.tex
\begin{figure}[t!]
\centering
\includegraphics[width=0.45\textwidth]{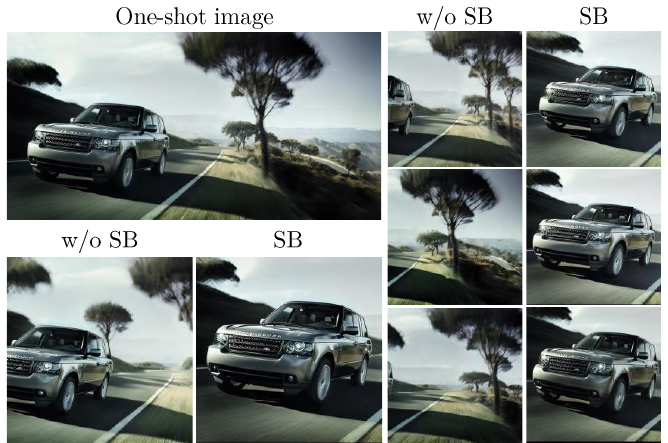}
\caption{
{\em Robust generation using semantic boosting (SB).} Without SB, a LoRA trained on a one-shot image often fails to render a car even when prompted with ``a photo of a car''. With SB, repeated exposure to the car region during training robustly generates complete cars.
}
\label{fig:sb}
\end{figure}

\begin{figure*}[t!]
\centering
\includegraphics[width=\textwidth]{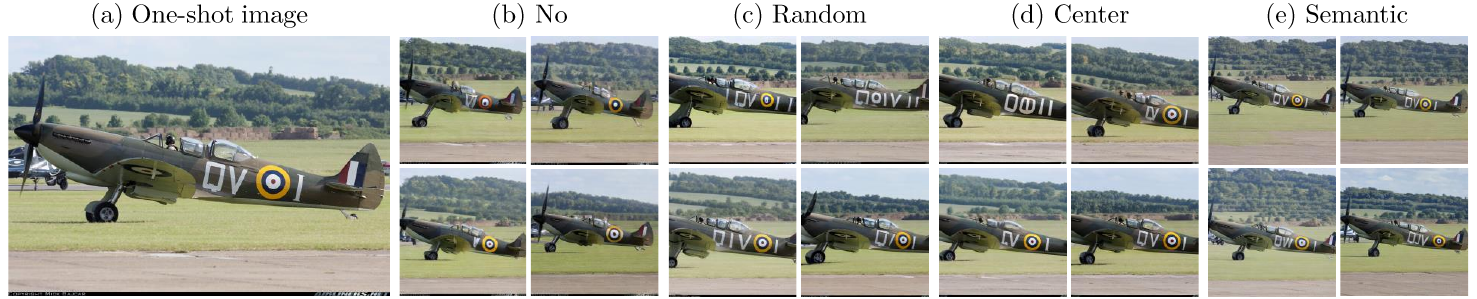}
\caption{{\em Effect of semantic boosting.} (a) The input image is used to train a LoRA under varying cropping methods. (b) Without cropping, the generated images exhibit a distorted aspect ratio of the primary object. (c, d) Conventional random and center cropping methods result in outputs where the object is consistently truncated. (e) In contrast, our semantic boosting preserves the object's structural integrity and details, leading to a robust generation.}
\vspace{-3mm}
\label{fig:sc}
\end{figure*}

\section{ChimeraLoRA}
To generate diverse and fine-grained synthetic images, we propose ChimeraLoRA.
We first describe a multi-head LoRA design trained with semantic boosting (Section~\ref{sec:alt}).
We then introduce a merging strategy that composes multiple heads for image generation (Section~\ref{sec:dir}).

\subsection{Multi-head LoRA Training}
\label{sec:alt}
Inspired by asymmetric LoRA architectures such as HydraLoRA~\citep{tian2024hydralora}, we separate the roles of LoRA into two parts: (i) a shared LoRA~$A$ that aggregates class-level knowledge and (ii) a set of image-wise LoRA heads $\mathcal{B} = \{ B_i \}^K_{i=1}$ that capture per-image details, as illustrated in Figure~\ref{fig:main}.
For simplicity, consider $K$ images $\{x_1, ..., x_K\}$ of a single label $y$.
We define a per-image reconstruction loss on image $x_i$ as follows:
\begin{align}
\!\!\! \mathcal{L}(A,B_i)
\! := \! \mathbb{E}_{\epsilon \sim \mathcal{N}(0,1),t}
\! \left[ \| \epsilon - \epsilon_{\theta,A,B_i}(z_{i,t}, t, \tau(y)) \|_2^2 \right]\!,
\label{eq:loss}
\end{align}
where $z_{i,t}$ is the noisy latent from an augmented view $f_{\text{aug}}(x_i)$ at time step $t$ and $\tau(y)$ denotes the class embedding from the pretrained text encoder $\tau$.
To capture the class prior, the shared LoRA~$A$ is optimized over total $K$ images by aggregating per-image objectives as follows:
\begin{align}
\mathcal{L}(A,\mathcal{B}) := \frac{1}{K} \sum_{i=1}^{K} \mathcal{L}(A,B_i) \;.
\label{eq:loss-A}
\end{align}
Following previous work~\citep{hu2022lora}, we initialize the LoRA~$A$ with random Gaussian weights and set each image-wise adapter $B_i$ to zero. We then jointly optimize $A$ and all $\{B_i\}_{i=1}^K$ by minimizing~\eqref{eq:loss-A}.
For stable training of the shared LoRA~$A$, we use distinct learning rates, setting LoRA~$A$’s rate lower than the~LoRA $B$'s ones~\citep{hayou2024lora+}.


Additionally, we focus on the noisy latent $z_{i,t}$ in~\eqref{eq:loss}. For robust training, practitioners typically employ data augmentation, and $z_{i,t}$ is therefore obtained from an augmented view $f_{\text{aug}}(x_i)$. However, common augmentations may not fully preserve the target class, which can be misaligned with the text prompt and can even hinder generating the target class, as illustrated in Figure~\ref{fig:sb}.

\paragraph{Semantic Boosting with Grounded-SAM.}
To emphasize class-level semantics shared across few-shot images, we propose a semantic boosting technique based on Grounded-SAM~\citep{liu2023grounding,kirillov2023segment}. Specifically, given an image $x$ with label $y$, we run a text-conditioned object detector using the text prompt for $y$ to produce candidate boxes and define $b^\star$ as the minimal enclosing box of the retained high-confidence targets. We then sample a crop region $\mathcal{R} \subset \mathbb{R}^2$ on $x$ with mild scaling and translation jitter, while enforcing $b^\star \subseteq \mathcal{R}$. 
To prevent the condition from being violated, we apply zero-padding to the original image, so that $b^\star$ remains fully visible and the crop meets the target size. This semantic cropping enables robust generation of the target class, as shown in Figure~\ref{fig:sb}. Furthermore, since our semantic cropping repeatedly exposes the target class region during training, the model better preserves the target’s aspect ratio and fine-grained details under the same training setup, as demonstrated in Figure~\ref{fig:sc}.

\subsection{LoRA Merging for Image Generation}
\label{sec:dir}
While each image-wise LoRA $B_i$ captures instance-specific details, generating synthetic images with a single adapter often fails to cover the full within-class distribution. Instead, we synthesize each image by combining the $K$ image-wise adapters with nonnegative weights sampled from a Dirichlet distribution as follows:
\begin{align}
B' = \sum_{i=1}^{K} w_i B_i,
\quad 
(w_1, \ldots, w_K) \sim \text{Dirichlet}(\boldsymbol{\alpha}) \;,
\label{eq:new-b}
\end{align}
where $\boldsymbol{\alpha} = \alpha\,\mathbf{1}_K = (\alpha, \ldots, \alpha)$.
After forming $B'$, we generate synthetic images by applying the class-adapter LoRA $A$ together with $B'$ to the base diffusion model.

In the symmetric Dirichlet case with $\boldsymbol{\alpha} \in \mathbb{R}^K$, the expectation and variance are computed as: 
\begin{align}
\mathbb{E}[w_i] = \frac{1}{K}, 
\quad
\mathrm{Var}[w_i] = \frac{K-1}{K^{2}\, (K\alpha+1)}\;.
\end{align}
Here, the concentration $\alpha$ controls how the mixture spreads over the simplex $\Delta^{K-1}$. When $\alpha=1$, $\mathbf{w}$ is uniformly distributed over the simplex. For $\alpha<1$, the distribution becomes sparse and typically concentrates most of its mass on a single $B_i$, which behaves like an image-wise regime. When $\alpha>1$, the weights cluster near the uniform vector and approximate a class-wise regime. We refer to this framework as ChimeraLoRA, which establishes a class-level backbone from the few-shot references and attaches instance-specific details to produce coherent semantics, yet diverse images.\footnote{In the Appendix~\ref{app:alpha-beta}, we expand our discussion to the setting with two concentration parameters, $\alpha$ and $\beta$, rather than a single $\alpha$.}


\paragraph{Remark.}
When $\alpha = 1$, $\mathbf{w}$ is uniform on the simplex $\Delta^{K-1}$, and we observe that this setting typically yields decent downstream performance. However, sampling a fresh $\mathbf{w}$ for every image can hinder batch-wise generation, as $B'$ must be rebuilt per sample. We consider two practical variants to mitigate this overhead: (i) set $w_i = 1/K$ in~\eqref{eq:new-b}, which still maintains high fidelity with reasonable coverage, as shown in Figure~\ref{fig:qual}; and (ii) reuse a single $\mathbf{w}\sim \text{Dirichlet}(\mathbf{1})$ to synthesize multiple images, with diffusion stochasticity providing additional variation. However, we note that per-image mixtures with $\text{Dirichlet}(\mathbf{1})$ still provide the broadest coverage. In Appendix~\ref{app:computation}, we analyze the trade-off between wall-time and accuracy across the three methods.

%% file: Sections/5_experiment.tex
\section{Experiment}
\label{sec:exp}

\subsection{Experimental Setup}
\paragraph{Datasets.}
We evaluate on 11 publicly available image classification datasets: FGVCAircraft (AIR)~\citep{maji2013fine}, Caltech101 (CAL)~\citep{fei2004learning}, StanfordCars (CAR)~\citep{krause20133d}, DTD~\citep{cimpoi2014describing}, EuroSAT (EUR)~\citep{helber2019eurosat}, Flowers102 (FLO)~\citep{nilsback2008automated}, Food101 (FOD)~\citep{bossard2014food}, OxfordPets (PET)~\citep{parkhi2012cats}, Skin Lesions (ISIC)~\citep{kassem2020skin}, CIFAR-10~\citep{krizhevsky2009learning}, and ImageNet100~\citep{russakovsky2015imagenet}. Our benchmarks cover diverse fine-grained tasks including cars and pets, and also specialized domains such as satellite imagery, textures, and medical dermatology, reflecting practical few-shot constraints in real applications.


\input{Figures_tex/t1_main}

\input{Figures_tex/t2_lt}

\paragraph{Implementation Details.}
We adopt CLIP ViT-B/16~\citep{dosovitskiy2021an,radford2021learning} as the downstream encoder. During fine-tuning, we attach rank-16 LoRA adapters to both the image and text encoders and train the model on synthetic training datasets derived from the given 4-shot references. Unless otherwise noted, all methods are trained for 60 epochs with AdamW~\citep{loshchilov2017decoupled} at a learning rate of $1\times10^{-4}$ using a cosine annealing scheduler. All experiments are run with three random seeds per setting, and we report the mean and variance.

\paragraph{Baselines.}
We compare our method against three methods: IsSynth~\citep{he2023is}, LoFT~\citep{kim2025loft}, and DataDream~\citep{kim2024datadream}. IsSynth is train-free and synthesizes data using features extracted from the given 4-shot references. LoFT and DataDream fine-tune diffusion models with LoRA in image-wise and class-wise configurations, respectively. To match trainable-parameter budgets in the 4-shot setting, DataDream uses LoRA rank 16, since it trains a single class-wise LoRA over all four images, whereas LoFT and our method use rank 4, as the adaption is split across the four images. We note that thanks to the shared adapter $A$, our approach uses 37.5\% fewer trainable parameters than both baselines. We employ a multi-head LoRA with distinct learning rates: $1 \times 10^{-4}$ for LoRA~$A$ and $1 \times 10^{-3}$ for LoRA~$B$.
All LoRA-based methods use Stable Diffusion 2.1~\citep{rombach2022high} as the base diffusion model. For generation, the trained adapters are attached to Stable Diffusion with guidance scale $2$, and when composing the per-image adapters we draw mixture coefficients from $\text{Dirichlet}(\mathbf{1})$. Unless otherwise noted, we reproduce the results for all baselines.


\subsection{Synthetic Datasets for Downstream Tasks}
\label{exp:few-shot}
\paragraph{Few-shot scenarios.}
We investigate whether synthetic datasets can surpass 4-shot real datasets. Table~\ref{tab:main-table} shows that fine-tuning CLIP with 4-shots per class attains an average accuracy of 71.8\% across nine datasets. We then generate 500 synthetic images per class and fine-tune CLIP on 504 images per class. In this setting, our ChimeraLoRA surpasses prior state-of-the-art methods. Notably, many baselines remain below the 4-shot real model even after adding synthetic data, underscoring a synthetic-to-real gap that limits practical utility.
Although our method also fails to surpass the 4-shot real model on DTD and FOD, its drop is smaller than competing approaches, with DataDream decreasing by 11.3 percentage points (pp) on EUR.
Overall, thanks to our synthetic images, ChimeraLoRA can build synthetic datasets that outperform real 4-shot datasets.



\paragraph{Long-tail scenarios.}
In practice, class frequencies are long-tailed rather than uniform with 4-shots per class. We therefore study augmenting only the tail classes with synthetic images. To simulate an extreme long-tailed regime, we split each dataset so that half of the classes are head classes with up to 500 real images and the other half are tail classes with 4-shots each. In this regime, training only on real images produces a model biased toward the long classes, as shown for EuroSAT in Table~\ref{tab:long-tail-table} where the accuracy gap between long and tail classes is 85.6 pp. Table~\ref{tab:long-tail-table} shows that adding synthetic images to tail classes leads to average accuracy gains, and our ChimeraLoRA outperforms baselines across five datasets. Especially, we observe that on ImageNet-100, adding synthetic images to the tail classes not only improves tail accuracy but also increases the accuracy of the long classes.

\begin{figure*}[t!]
\centering
\includegraphics[width=\textwidth]{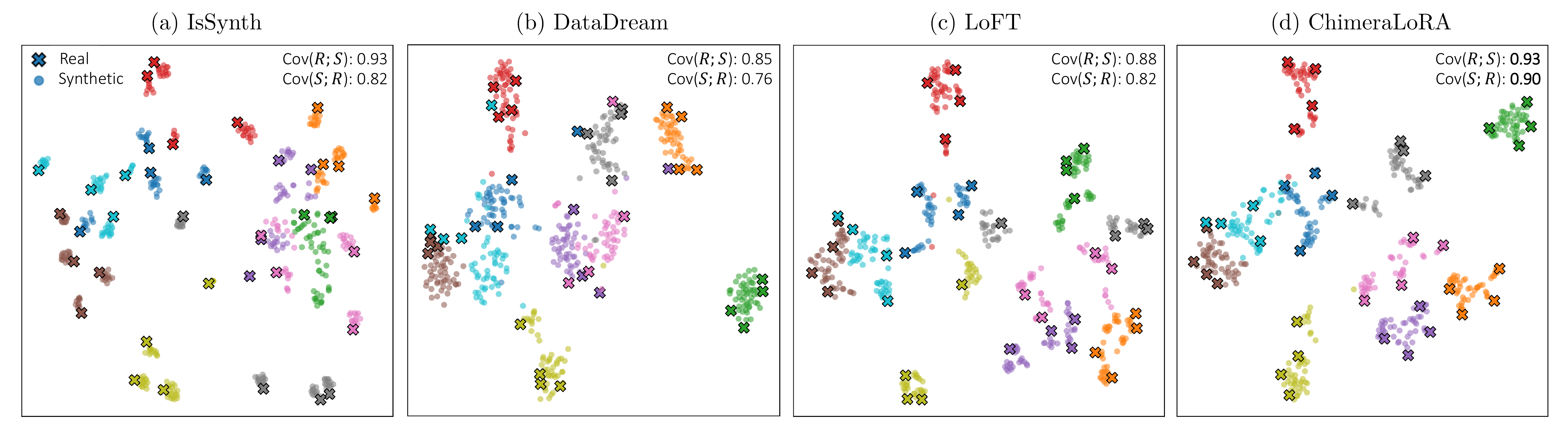}
\caption{{\em t-SNE for real and synthetic images.} ChimeraLoRA generates mainly inside the region spanned by the real anchors marked with crosses and attains the highest coverage across methods, with Cov$(\mathcal{R};\mathcal{S}) = 0.93$ and Cov$(\mathcal{S};\mathcal{R}) = 0.90$.}
\label{fig:tsne-dataset}
\vspace{-3mm}
\end{figure*}

\begin{figure}[t!]
\centering
\includegraphics[width=0.45\textwidth]{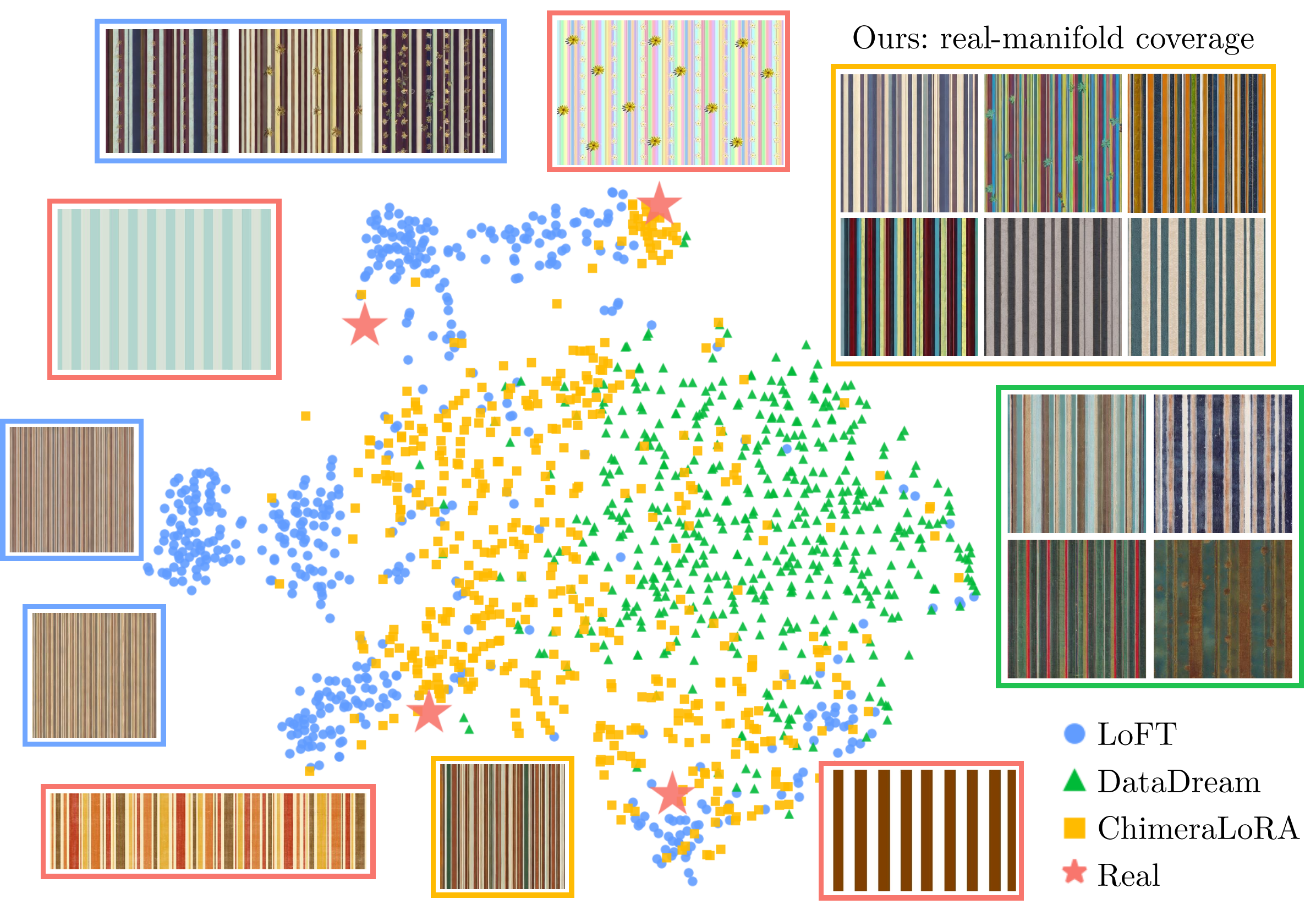}
\caption{
{\em Real-manifold coverage.} Our ChimeraLoRA samples (yellow rectangles) fall within the region spanned by the four real anchors (red stars), indicating coverage of the real manifold, whereas the baselines drift outside.
}
\label{fig:tsne-class}
\vspace{-2mm}
\end{figure}


\subsection{Synthetic-to-Real Gap Analyses}
\label{sec:analyses}
\paragraph{Within-class visualization.}
In Section~\ref{sec:dir}, we merge multiple LoRA heads with weights sampled from Dirichlet$(\mathbf{1})$ and use the merged adapter for image synthesis. Figure~\ref{fig:tsne-class} shows that such Dirichlet-weighted mixtures, which lie inside the probability simplex, transfer to actual generations. Specifically, we visualize the \textit{banded} class from DTD by taking four real images and, for each method, 500 synthetic images, for a total of 1,504 images.
Then, we compute CLIP image embeddings, reduce dimensionality with PCA~\cite{shlens2014tutorial}, and apply t-SNE~\citep{maaten2008visualizing}. 
Here, red stars mark the real anchors.
As intended, ChimeraLoRA spreads its samples roughly uniformly within the real region, whereas the baselines place many samples outside it.
In addition, LoFT collapses into a few tight clusters, producing near-duplicates, while DataDream drifts farther from the anchors and often yields lower-fidelity samples with wavy banding and inconsistent colors. For clarity, we recommend zooming in on Figure~\ref{fig:tsne-class}.

\paragraph{Cross-class visualization.}
Beyond the single-class analysis, Figure~\ref{fig:tsne-dataset} presents a cross-class visualization of ten DTD classes.
We use 4 real images and 50 synthetic images per class, for 540 images in total. Similarly to the single-class case, we compute CLIP image embeddings and visualize them with t-SNE. 
To evaluate how well the real and synthetic sets cover one another, we report two directional coverages. Let the real set be $\mathcal{R}$ and the synthetic set be $\mathcal{S}$. Let $\phi(\cdot)$ be the L2-normalized CLIP image embedding and define the cosine distance $\delta(\mathbf{u},\mathbf{v}) = 1 - \langle \mathbf{u}, \mathbf{v} \rangle$. Define a class radius $\rho$ from the median real-real nearest-neighbor distance in CLIP space. With $\mathcal{R}$ as the anchor, the coverage of $\mathcal{R}$ by $\mathcal{S}$ is defined as follows:
\begin{align}
\!\!\!\! \text{Cov}(\mathcal{R}; \mathcal{S}) \! = \! \frac{1}{|\mathcal{R}|}\! \sum_{r \in \mathcal{R}} \! \mathbf{1} \!\left[\exists s\in \mathcal{S}\ \text{s.t.}\ \delta \big(\phi(r),\phi(s)\big)\le \rho\right]\!.\!
\end{align}
The symmetric measure $\mathrm{Cov}(\mathcal{S}; \mathcal{R})$ is defined by swapping $\mathcal{R}$ and $\mathcal{S}$. Figure~\ref{fig:tsne-dataset} demonstrates that ChimeraLoRA attains higher scores on both measures than other methods, indicating that real and synthetic images intermix more naturally. 

\input{Figures_tex/t4_fid_clip}

\paragraph{Quantifying the synthetic-to-real gap.} 
We evaluate the synthetic-to-real gap relative to the 4-shots real reference using three metrics computed per class and averaged across classes. First, Fréchet Inception Distance (FID) is computed in CLIP image-embedding space rather than Inception~\citep{szegedy2017inception}, following recent work~\citep{jayasumana2024rethinking}. As a closer distributional match yields a smaller FID, the Real 4-shot row has FID = 0 in Table~\ref{tab:fd}. Second, CLIP score is the cosine similarity between each image’s CLIP embedding and the class text embedding, reported after multiplying by 100. Third, centroid similarity is the cosine similarity in CLIP space between the centroid of 500 synthetic images and the centroid of the 4 real images for each class, normalized so that the Real 4-shots row equals 100.0. Table~\ref{tab:fd} shows that ChimeraLoRA attains lower FID and higher CLIP score and centroid similarity than the baselines, indicating that our method exhibits the smallest synthetic-to-real gap.


\subsection{Further Ablation Studies}
We conduct a range of ablation studies on our method, and additional ablation results are provided in the Appendix~\ref{app:ablation}.

\vspace{-2mm}
\paragraph{Ablation on the number of synthetic images.}
As generating 500 synthetic images per class is costly, we conduct experiments on smaller number of synthetic images. Figure~\ref{fig:scale} shows that ChimeraLoRA maintains robust performance, and its accuracy increases with the number of synthetic images as the added synthetic images align with the real distribution induced by the few-shot references. However, on the EUR and AIR datasets, DataDream and LoFT exhibit declining accuracy as more synthetic images are added, revealing a persistent synthetic-to-real gap that worsens with scale.


\input{Figures_tex/g1_scale}

\begin{figure}[t!]
\centering
\includegraphics[width=0.47\textwidth]{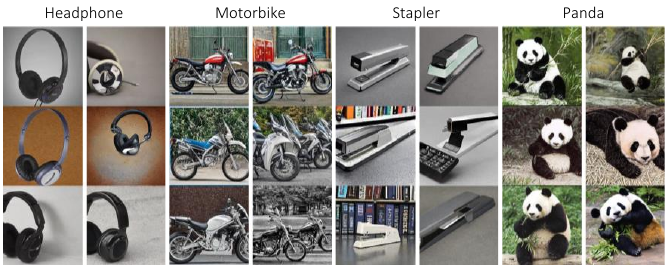}
\caption{
{\em Synthetic images with sharing LoRA $B$.} For each class, the left three images are from our ChimeraLoRA with a shared LoRA~$A$, and the right three are from a variant with a shared LoRA~$B$. The right images look more diverse but often miss the target object or fine details such as a motorcycle’s wheel.
}
\label{fig:multi}
\vspace{-3mm}
\end{figure}

\paragraph{Ablation on shared LoRA parts.}
In Section~\ref{sec:alt} we introduce a shared LoRA~$A$ to encode class priors instead of sharing LoRA~$B$. Figure~\ref{fig:multi} compares the results of shared LoRA~$A$ on the left and shared LoRA~$B$ on the right for each class. With shared $B$, images often look more diverse yet miss the target object or fine details, such as truncated headphone ends, awkward staplers, and motorbikes lacking inner wheel detail. 
Conceptually, LoRA~$A$ plays an encoder-like role that projects features into a shared rank-$r$ subspace, while LoRA~$B$ acts as a decoder that lifts this representation back to the full model space. 
When the few shot references share the same semantics, sharing the encoder $A$ promotes a class consistent encoding, and instance specific decoders $B$ can then reconstruct high frequency structure. This division yields better object integrity and sharper details than sharing $B$. Our observation aligns with asymmetric roles reported in previous work~\citep{pmlr-v235-zhu24c,guo2025selective}: LoRA~$B$ is more tightly coupled to the input data distribution, whereas LoRA~$A$ aggregates shared knowledge.

\vspace{-2mm}
\paragraph{Ablation on proposed components.}
Table~\ref{tab:comp} isolates the effects of multi-head LoRA and semantic boosting. Removing both components reduces our method to LoFT~\citep{kim2025loft}, which uses only per-image LoRAs without a class-shared adapter or box guidance and yields the lowest accuracy. Introducing multi-head LoRA brings a class-shared adapter that coordinates the per-image adapters and delivers clear gains. Incorporating semantic boosting preserves class bounding boxes during fine-tuning and further improves accuracy. Neither component is uniformly superior across datasets. Combined in ChimeraLoRA, their benefits compound and the model attains the best overall results.


\input{Figures_tex/t3_component}





%% file: Figures_tex/t1_main.tex
\begin{table*}[!t]
\setlength\tabcolsep{4pt}
\centering
\small
\caption{{\em Downstream performance with synthetic datasets under 4-shot scenarios.} Starting with 4-shot labels, we generate 500 additional images per class to train on 504 per class in total, improving accuracy by 2.1 percentage points on average over state-of-the-art baselines across nine datasets. We mark the best in bold and the second best with underlines.}
\begin{tabular}{l|ccccccccc|c}
\toprule
\textbf{Methods} & \textbf{AIR} & \textbf{CAL} & \textbf{CAR} & \textbf{DTD} & \textbf{EUR} & \textbf{FLO} & \textbf{FOD} & \textbf{PET} & \textbf{ISIC} & \textbf{AVG} \\

\midrule    
CLIP (0-shot)
& 24.7 & 93.0 & 65.2 & 44.4 & 47.6 & 71.4 & 86.1 & 89.2 & 21.1 & 60.3 \\
CLIP (4-shots) 
& $41.3_{\pm 0.3}$ & $95.5_{\pm 0.1}$ & $74.3_{\pm 0.2}$ & $62.0_{\pm 0.7}$ & $83.5_{\pm 0.6}$ & $89.9_{\pm 0.4}$ & $86.5_{\pm 0.1}$ & $93.3_{\pm 0.4}$ & $19.6_{\pm 1.5}$ & $71.8_{\pm 0.4}$ \\
\midrule
IsSynth~\citep{he2023is}
& $39.9_{\pm 0.1}$ & $95.5_{\pm 0.1}$ & $71.5_{\pm 0.6}$ & \underline{60.1}$_{\pm 0.3}$ & $73.4_{\pm 0.6}$ & $89.0_{\pm 1.0}$ & $85.6_{\pm 0.1}$ & $91.6_{\pm 0.2}$ & $23.8_{\pm 1.4}$ & $70.1_{\pm 0.4}$ \\
DataDream~\citep{kim2024datadream}
& \underline{44.3}$_{\pm 0.4}$ & \textbf{96.1}$_{\pm 0.1}$ & \textbf{81.7}$_{\pm 0.3}$ & $56.0_{\pm 0.6}$ & $72.2_{\pm 0.7}$ & \underline{92.9}$_{\pm 0.5}$ & \textbf{86.0}$_{\pm 0.1}$ & $92.2_{\pm 0.1}$ & $20.7_{\pm 1.1}$ & $71.3_{\pm 0.3}$ \\
LoFT~\citep{kim2025loft}
& $41.7_{\pm 0.6}$ & $95.7_{\pm 0.1}$ & $78.0_{\pm 0.3}$ & $58.0_{\pm 1.5}$ & \underline{85.0}$_{\pm 0.7}$ & $91.3_{\pm 0.2}$ & $85.1_{\pm 0.1}$ & \underline{92.4}$_{\pm 0.3}$ & \underline{25.6}$_{\pm 0.3}$ & \underline{72.5}$_{\pm 0.4}$ \\
\rowcolor{gray!15}
ChimeraLoRA
& \textbf{46.0}$_{\pm 0.7}$ & \textbf{96.1}$_{\pm 0.1}$ & \underline{79.6}$_{\pm 0.5}$ & \textbf{61.6}$_{\pm 0.5}$ & \textbf{86.3}$_{\pm 0.5}$ & \textbf{93.4}$_{\pm 0.4}$ & \underline{85.7}$_{\pm 0.1}$ & $\bm{93.4}_{\pm 0.1}$ & $\bm{29.2}_{\pm 0.6}$ & \textbf{74.6}$_{\pm 0.2}$ \\
\bottomrule
\end{tabular}
\vspace{2mm}
\label{tab:main-table}
\end{table*}

%% file: Figures_tex/t2_lt.tex
\begin{table*}[!t]
\setlength\tabcolsep{4pt}
\centering
\small
\caption{{\em Downstream performance with synthetic datasets under long-tail scenarios.} When training with only 4 samples per tail class, the classifier’s decision boundary skews toward the long classes, yielding poor tail performance. After adding 500 synthetic images per tail class with ChimeraLoRA, accuracy improves by 7.62 percentage points on average relative to the real-only baseline, with a 14.74 percentage-point gain on the tail classes specifically.}
\begin{tabular}{l|ccc|ccc|ccc|ccc|ccc}
\toprule
\multirow[c]{2}{*}{\raisebox{-0.3em}{\textbf{Methods}}}
  & \multicolumn{3}{c|}{\textbf{CIFAR10}}
  & \multicolumn{3}{c|}{\textbf{ImageNet100}}
  & \multicolumn{3}{c|}{\textbf{DTD}} 
  & \multicolumn{3}{c|}{\textbf{EuroSAT}}
  & \multicolumn{3}{c}{\textbf{Flowers102}} \\
\cmidrule(lr){2-4}\cmidrule(lr){5-7}\cmidrule(lr){8-10}\cmidrule(lr){11-13}\cmidrule(lr){14-16}
& \textbf{Head} & \textbf{Tail} & \textbf{Avg.} & \textbf{Head} & \textbf{Tail} & \textbf{Avg.} & \textbf{Head} & \textbf{Tail} & \textbf{Avg.} & \textbf{Head} & \textbf{Tail} & \textbf{Avg.} & \textbf{Head} & \textbf{Tail} & \textbf{Avg.} \\
\midrule
Real & \textbf{98.8} & 70.1 & 84.5 & 78.8 & 79.1 & 79.0 & \textbf{86.5} & 46.7 & 66.6 & \textbf{99.2} & 13.6 & 56.4 & 92.8 & 94.2 & 93.5 \\
DataDream~\citep{kim2024datadream} & 98.1 & 76.3 & 87.2 & 88.7 & \underline{91.3} & \underline{90.0} & 78.8 & \underline{55.5} & 67.2 & \underline{98.8} & \underline{48.7} & \underline{73.9} & \underline{93.3} & 95.5 & \underline{94.4} \\
LoFT~\citep{kim2025loft} & \underline{98.4} & \underline{76.5} & \underline{87.4} & \textbf{88.9} & 91.2 & \underline{90.0} & \underline{84.3} & 51.6 & \underline{67.9} & 98.7 & 47.2 & 73.0 & 91.5 & \underline{96.0} & 93.7 \\
\rowcolor{gray!15}
ChimeraLoRA & 98.3 & \textbf{81.0} & \textbf{89.6} & \underline{88.8} & \textbf{91.6} & \textbf{90.2} & 80.2 & \textbf{56.6} & \textbf{68.4} & 98.0 & \textbf{51.3} & \textbf{74.5} & \textbf{93.9} & \textbf{96.9} & \textbf{95.4} \\
\bottomrule
\end{tabular}
\vspace{1mm}
\label{tab:long-tail-table}
\end{table*}

%% file: Figures_tex/t4_fid_clip.tex
\begin{table}[!t]
\centering
\small
\caption{{\em Synthetic-to-real gap analyses.} Across nine datasets, ChimeraLoRA produces synthetic images that most closely match the 4-shot real reference on average, with the lowest FID@4 and the highest CLIP score and centroid similarity.}
\centering
{\setlength\tabcolsep{3pt}
\begin{tabular}{l|ccc}
\toprule
\textbf{Methods} & \textbf{FID@4} $\downarrow$ & \textbf{CLIP score} $\uparrow$ & \textbf{Centroid Sim.} $\uparrow$ \\
\midrule
Real (4-shots) & 0.00 & 29.48 & 100.0 \\
\midrule
DataDream~\citep{kim2024datadream} & 0.23 & 29.67 & 87.8 \\
LoFT~\citep{kim2025loft} & 0.22 & 30.04 & 90.1 \\ 
\rowcolor{gray!15}
ChimeraLoRA & \textbf{0.20} & \textbf{30.31} & \textbf{90.5} \\
\bottomrule
\end{tabular}
}
\vspace{-2mm}
\label{tab:fd}
\end{table}

%% file: Figures_tex/g1_scale.tex
\begin{figure}[!t]
    \centering
    \hspace{-5mm}
    \begin{subfigure}[t]{0.45\linewidth}
        \centering
        \begin{tikzpicture}
            \begin{axis}[
                scale only axis,
                enlargelimits=false,
                clip mode=individual,
                label style={font=\scriptsize},
                tick label style={font=\scriptsize},
                width=0.85\linewidth,
                height=0.89\linewidth,
                xlabel=\# of synthetic images per class,
                ylabel=Accuracy (\%),
                xmin=-25, xmax=550,
                ymin=71, ymax=88,
                xtick={25, 100, 250, 500},
                ytick={75, 80, 85},
                xlabel style={yshift=0.5em},
                ylabel style={yshift=-1.8em},
                mark size=1.4pt,
                legend style={
                    nodes={scale=0.6}, 
                    at={(2.1, 1.2)}, 
                    /tikz/every even column/.append style={column sep=1mm},
                    legend columns=4,
                    legend image post style={mark size=2.4pt, xscale=0.5, yscale=0.5}
                },
            ]
            
            \addplot[cRed, semithick, mark=triangle*, mark options={solid}] table[col sep=comma, x=x, y=euro_ours]{Data/scale.csv};
            \addplot[orange, semithick, mark=square*, mark options={solid}] table[col sep=comma, x=x, y=euro_loft]{Data/scale.csv};
            \addplot[cGreen, semithick, mark=pentagon*, mark options={solid}] table[col sep=comma, x=x, y=euro_datadream]{Data/scale.csv};
            \addplot[cBlue, semithick, mark=*, mark options={solid}] table[col sep=comma, x=x, y=euro_issynth]{Data/scale.csv};

            \addplot[name path=ours_u, draw=none, fill=none] table[col sep=comma, x=x, y=euro_ours_u]{Data/scale.csv};
            \addplot[name path=ours_d, draw=none, fill=none] table[col sep=comma, x=x, y=euro_ours_d]{Data/scale.csv};
            \addplot[cRed, fill opacity=0.15] fill between[of=ours_u and ours_d];

            \addplot[name path=loft_u, draw=none, fill=none] table[col sep=comma, x=x, y=euro_loft_u]{Data/scale.csv};
            \addplot[name path=loft_d, draw=none, fill=none] table[col sep=comma, x=x, y=euro_loft_d]{Data/scale.csv};
            \addplot[orange, fill opacity=0.15] fill between[of=loft_u and loft_d];

            \addplot[name path=datadream_u, draw=none, fill=none] table[col sep=comma, x=x, y=euro_datadream_u]{Data/scale.csv};
            \addplot[name path=datadream_d, draw=none, fill=none] table[col sep=comma, x=x, y=euro_datadream_d]{Data/scale.csv};
            \addplot[cGreen, fill opacity=0.15] fill between[of=datadream_u and datadream_d];

            \addplot[name path=issynth_u, draw=none, fill=none] table[col sep=comma, x=x, y=euro_issynth_u]{Data/scale.csv};
            \addplot[name path=issynth_d, draw=none, fill=none] table[col sep=comma, x=x, y=euro_issynth_d]{Data/scale.csv};
            \addplot[cBlue, fill opacity=0.15] fill between[of=issynth_u and issynth_d];
            
            \legend{ChimeraLoRA (ours), LoFT, DataDream, IsSynth}
            
            \end{axis}
        \end{tikzpicture}
        \caption{EuroSAT}
    \end{subfigure}
    \hspace{2mm}
    \begin{subfigure}[t]{0.45\linewidth}
        \centering
        \begin{tikzpicture}
            \begin{axis}[
                scale only axis,
                enlargelimits=false,
                clip mode=individual,
                label style={font=\scriptsize},
                tick label style={font=\scriptsize},
                width=0.85\linewidth,
                height=0.89\linewidth,
                xlabel=\# of synthetic images per class,
                ylabel=Accuracy (\%),
                xmin=-25, xmax=550,
                ymin=39, ymax=47,
                xtick={25, 100, 250, 500},
                ytick={39, 42, 45},
                xlabel style={yshift=0.5em},
                ylabel style={yshift=-1.8em},
                mark size=1.4pt,
                legend style={
                    nodes={scale=0.6}, 
                    at={(0.97, 0.2)}, 
                    /tikz/every even column/.append style={column sep=1mm},
                    legend columns=2,
                    legend image post style={mark size=2.4pt, xscale=0.5, yscale=0.5}
                },
            ]

            \addplot[cRed, semithick, mark=triangle*, mark options={solid}] table[col sep=comma, x=x, y=air_ours]{Data/scale.csv};
            \addplot[orange, semithick, mark=square*, mark options={solid}] table[col sep=comma, x=x, y=air_loft]{Data/scale.csv};
            \addplot[cGreen, semithick, mark=pentagon*, mark options={solid}] table[col sep=comma, x=x, y=air_datadream]{Data/scale.csv};
            \addplot[cBlue, semithick, mark=*, mark options={solid}] table[col sep=comma, x=x, y=air_issynth]{Data/scale.csv};

            \addplot[name path=ours_u, draw=none, fill=none] table[col sep=comma, x=x, y=air_ours_u]{Data/scale.csv};
            \addplot[name path=ours_d, draw=none, fill=none] table[col sep=comma, x=x, y=air_ours_d]{Data/scale.csv};
            \addplot[cRed, fill opacity=0.15] fill between[of=ours_u and ours_d];

            \addplot[name path=loft_u, draw=none, fill=none] table[col sep=comma, x=x, y=air_loft_u]{Data/scale.csv};
            \addplot[name path=loft_d, draw=none, fill=none] table[col sep=comma, x=x, y=air_loft_d]{Data/scale.csv};
            \addplot[orange, fill opacity=0.15] fill between[of=loft_u and loft_d];

            \addplot[name path=datadream_u, draw=none, fill=none] table[col sep=comma, x=x, y=air_datadream_u]{Data/scale.csv};
            \addplot[name path=datadream_d, draw=none, fill=none] table[col sep=comma, x=x, y=air_datadream_d]{Data/scale.csv};
            \addplot[cGreen, fill opacity=0.15] fill between[of=datadream_u and datadream_d];

            \addplot[name path=issynth_u, draw=none, fill=none] table[col sep=comma, x=x, y=air_issynth_u]{Data/scale.csv};
            \addplot[name path=issynth_d, draw=none, fill=none] table[col sep=comma, x=x, y=air_issynth_d]{Data/scale.csv};
            \addplot[cBlue, fill opacity=0.15] fill between[of=issynth_u and issynth_d];

            \end{axis}
        \end{tikzpicture}
        \caption{FGVCAircraft}
    \end{subfigure}
\vspace{-2mm}
\caption{{\em Robustness under scaling the synthetic budget.} Accuracy rises with more synthetic images per class for ChimeraLoRA. Shaded regions indicate variability across seeds. }
\label{fig:scale}
\vspace{-3mm}
\end{figure}

%% file: Figures_tex/t3_component.tex
\begin{table}[!t]
\centering
\small
\caption{{\em Component ablation.} Each proposed components contribute to performance gains.}
\centering
{\setlength\tabcolsep{4pt}
\begin{tabular}{cc|cc}
\toprule
\textbf{Multi-Head LoRA} & \textbf{Semantic Boosting} & \textbf{AIR} & \textbf{FLO} \\
\midrule
\xmark & \xmark & 41.7 & 91.3 \\
\cmark & \xmark & 43.9 & 93.1 \\
\xmark & \cmark & 44.4 & 92.2 \\
\rowcolor{gray!15}
\cmark & \cmark & \textbf{46.0} & \textbf{93.4} \\
\bottomrule
\end{tabular}
}
\label{tab:comp}
\vspace{-3mm}
\end{table}

%% file: Sections/6_conclusion.tex
\section{Conclusion}
In this work, we propose a multi-head LoRA guided method for synthetic dataset generation, called ChimeraLoRA. By separating the roles of two low-rank adapters, we use a class-shared adapter to encode class priors and per-image adapters to model instance-level details. To help the shared adapter capture class semantics, we introduce semantic boosting that leverages class bounding boxes during adapter fine-tuning. For image synthesis, we fix the class-shared adapter and merge the per-image adapters to generate images with high diversity and fidelity, which in turn improve downstream task performance. Extensive experiments across diverse classification tasks and practical domains, including medical applications and long-tailed scenarios, show that our method outperforms baselines.

\paragraph{Limitations and future work.}
We use Grounded-SAM as a general solution across domains for Semantic Boosting, but in medical settings, domain-specific tools such as MedSAM may be more appropriate, and our current medical domain validation remains limited. Future work will strengthen domain-specific evidence by evaluating on additional medical datasets, conducting robustness analyses under clinically relevant perturbations, and expanding the discussion of related work on generative augmentation for long-tailed medical image classification rather than assuming that simple tool substitution is sufficient.

\paragraph{Acknowledgements.}
This work was partly supported by the IITP grants and the NRF grants funded by Ministry of Science and ICT, Korea (No.RS-2019-II191906, Artificial Intelligence Graduate School Program
(POSTECH); No.RS-2024-00457882, AI Research Hub Project; IITP-2026-RS-2024-00437866; RS-2024-00509258, Global AI Frontier Lab).

%% file: Sections/X_suppl.tex
\clearpage
\setcounter{page}{1}
\maketitlesupplementary

\renewcommand{\thesection}{\Alph{section}}
\renewcommand{\thesubsection}{\thesection.\arabic{subsection}}
\setcounter{section}{0} 

\begin{strip}
\centering
\setlength\tabcolsep{4pt}
\small
\begin{tabular}{l|ccccccccc|c}
\toprule
{Datasets} & {AIR} & {CAL} & {CAR} & {DTD} & {EUR} & {FLO} & {FOD} & {PET} & \textbf{SUN} & {AVG} \\
\midrule    
IsSynth
& 34.5 & 95.7 & 65.9 & 67.6 & 71.3 & 90.0 & 85.4 & 92.1 & 72.2 & 75.0 \\
$\text{DataDream}$ 
& 61.4 & 96.0 & \underline{90.5} & 64.9 & 84.1 & \underline{98.0} & \underline{86.5} & \underline{93.5} & 74.4 & 83.3 \\
LoFT
& \underline{66.1} & \underline{96.7} & 89.3 & \underline{70.5} & \underline{86.8} & \underline{98.0} & 86.0 & 93.2 & \underline{75.5} & \underline{84.7} \\
\rowcolor{Gray} Ours
& \textbf{71.4} & \textbf{97.1} & \textbf{90.8} & \textbf{74.7} & \textbf{92.3} & \textbf{98.6} & \textbf{87.1} & \textbf{94.2} & \textbf{76.6} & \textbf{87.0} \\
\bottomrule
\end{tabular}
\captionof{table}{{\em Downstream performance with synthetic datasets under 16-shot scenarios.} Starting with 16-shot labels, we generate 500 additional images per class to train on 504 per class in total, improving accuracy by 2.3 percentage points on average over state-of-the-art baselines across nine datasets. Here, the baseline results are taken from the original papers.}
\label{app:main-table}
\end{strip}

\section{Experiments on 16-Shot Setting}
\label{app:alpha-beta}

\noindent
Table~\ref{app:main-table} shows that our method works well even in the 16-shot setting. Rather than using a single parameter $\alpha$, we employ two concentration parameters, $\alpha$ and $\beta$, to define the Dirichlet distribution. Specifically, we sample weights $(w_1,\dots,w_{16})\sim \mathrm{Dir}(\mathbf{c})$, where $\mathbf{c}\in\mathbb{R}^{16}$ is defined as:
\begin{align}
c_i = 
\begin{cases}
\alpha & \text{if } i=k \\
\beta & \text{if } i \neq k
\end{cases},
\quad
k \sim \text{Uniform}(\{1, \dots, 16\}) \;.
\end{align}
\vspace{-4mm}

\noindent
We set $\alpha=2$ and $\beta=0.07$, yielding $\mathbb{E}[w_k]=\frac{\alpha}{\alpha+(16-1)\beta} \approx 66\%$ for the main adapter, with the remaining $\approx 34\%$ mass distributed over the other adapters. Choosing a small $\beta<1$ promotes sparsity among the remaining adapters, i.e., the residual mass is typically concentrated on only a few adapters rather than being spread uniformly across all of them.
In summary, as the number of shots increases, using a relatively larger $\alpha$ (relative to $\beta$) becomes beneficial for capturing fine-grained details by placing more emphasis on the main adapter, while still preventing too many adapters from being mixed in.

\section{Computation Analyses}
\subsection{Resource Usage}
As each image has its own LoRA head, resource usage naturally grows as the number of shots increases, leading to higher memory consumption and more trainable parameters. To keep this overhead manageable, we reduce the LoRA rank for multi-shot settings. In particular, the results reported in Table~\ref{tab:main-table} for the 4-shot setting are obtained by setting the rank to one quarter of the baseline, which substantially lowers the per-head footprint while retaining competitive performance. Table~\ref{tab:resource} further summarizes the resulting parameter counts and training time, showing that our approach offers a practical middle ground between baselines in terms of both memory and speed.

\begin{table}[!ht]
\centering
\setlength\tabcolsep{4pt}
\small
\begin{tabular}{l|ccc}
\toprule
Method & Rank & \# parameters & Training time \\
\midrule
DataDream & 16 & 3.2M & \textbf{95s} \\
LoFT      & 4  & \textbf{0.8M} & 356s \\
Ours      & 4  & \underline{1.9M} & \underline{212s} \\
\bottomrule
\end{tabular}
\caption{Trade-off between \# parameters and training time.}
\label{tab:resource}
\end{table}
    
\subsection{Effect of Weight Sampling on Efficiency}
\label{app:computation}
In Section~\ref{sec:dir}, when $\alpha=1$ we draw $\mathbf{w}\sim\mathrm{Dirichlet}(\mathbf{1})$, i.e., $\mathbf{w}$ is uniformly distributed over the simplex $\Delta^{K-1}$, and we compare three practical choices: \emph{Uniform $w$} (fix $w_i=1/K$), \emph{Reuse $w$} (reuse a single $\mathbf{w}\sim\mathrm{Dirichlet}(\mathbf{1})$ for multiple images), and \emph{New $w$} (sample a fresh $\mathbf{w}$ per image). Table~\ref{tab:wall-time} shows that moving from Uniform $\rightarrow$ Reuse $\rightarrow$ New increases the diversity of per-image mixtures and tends to improve accuracy, but also increases wall-time, where wall-time denotes the overall generation time, since per-image sampling reduces batching efficiency.

\begin{table}[!h]
\centering
\setlength\tabcolsep{4pt}
\small
\begin{tabular}{l|cc}
\toprule
Method & Wall-Time & Accuracy \\
\midrule
Uniform $w$ & \textbf{4.6h} & 59.37 \\
Reuse $w$ & \underline{5.8h} & \underline{61.31} \\
New $w$ & 6.1h & \textbf{61.51} \\
\bottomrule
\end{tabular}
\caption{Trade-off between wall-time and accuracy.}
\label{tab:wall-time}
\end{table}

\section{Additional Ablation Studies}
\label{app:ablation}

\subsection{Rank \texorpdfstring{$r$}{r} Ablation}
On the DTD dataset, we observe that increasing the LoRA rank $r$ yields modest but consistent gains: with the number of synthetic images fixed at 100, performance improves from 55.07 at $r{=}2$ to 57.20 and 57.72 at $r{=}8$ and $r{=}16$, respectively. This trend aligns with the increased capacity of higher-rank adapters, but it also comes with higher computational cost, as larger $r$ increases the number of trainable parameters and the associated memory and training-time overhead. Overall, we find that higher ranks provide incremental improvements, reflecting a typical trade-off between computation resources and performance.

\subsection{\texorpdfstring{$\alpha$}{alpha} Ablation}
In a symmetric Dirichlet distribution, \texorpdfstring{$\alpha$}{alpha} primarily controls the concentration of the mixture weights $\mathbf{w}$: smaller $\alpha$ produces spikier weights that are closer to selecting a single head, whereas larger $\alpha$ spreads mass more evenly and behaves like a more uniform ensemble. As summarized in Table~\ref{tab:alpha}, this induces a clear trade-off. With smaller \texorpdfstring{$\alpha$}{alpha}, synthesis tends to be more faithful and yields better FID, while larger $\alpha$ encourages broader mixing across heads and improves diversity, reflected by a higher distance score. Overall, $\alpha$ serves as a simple tunable parameter to balance fidelity and diversity depending on the desired operating point.

\begin{table}[!h]
\centering
\small
\label{tab:alpha_metrics}
\begin{tabular}{lccc}
\toprule
Metric & $\alpha=0.2$ & $\alpha=1.0$ & $\alpha=5.0$ \\
\midrule
FID (Real vs. Synthetic, $\downarrow$)                 & \textbf{0.098} & \underline{0.101} & 0.146 \\
Distance (Inter-class, $\uparrow$)  & 0.142 & \underline{0.153} & \textbf{0.199} \\
\bottomrule
\end{tabular}
\caption{{\em Comparison metrics under different $\alpha$ values.}}
\label{tab:alpha}
\end{table}

\subsection{Semantic Boosting on Baselines}
The table~\ref{tab:sb} summarizes the effect of applying semantic boosting to the baselines. Overall, semantic boosting consistently improves performance, and the gain is most pronounced for image-wise LoRA training, where optimization can otherwise overfit to spurious regions or drift away from the target instance.

\begin{table}[!ht]
\centering
\small
\begin{tabular}{l|cc}
\toprule
Method & AIR & FLO \\
\midrule
DataDream (+SB) & 44.3 $\rightarrow$ 44.2 & 92.9 $\rightarrow$ \underline{93.1} \\
LoFT (+SB)     & 41.7 $\rightarrow$ \underline{44.4} & 91.3 $\rightarrow$ 92.2 \\
\rowcolor{Gray}
Ours & 43.9 $\rightarrow$ \textbf{46.0} & 93.1 $\rightarrow$ \textbf{93.4} \\
\bottomrule
\end{tabular}
\caption{{\em Effect of semantic boosting on baselines.}}
\label{tab:sb}
\end{table}

\section{Additional Qualitative Results}
\begin{figure*}[t!]
\centering
\includegraphics[width=0.8\textwidth]{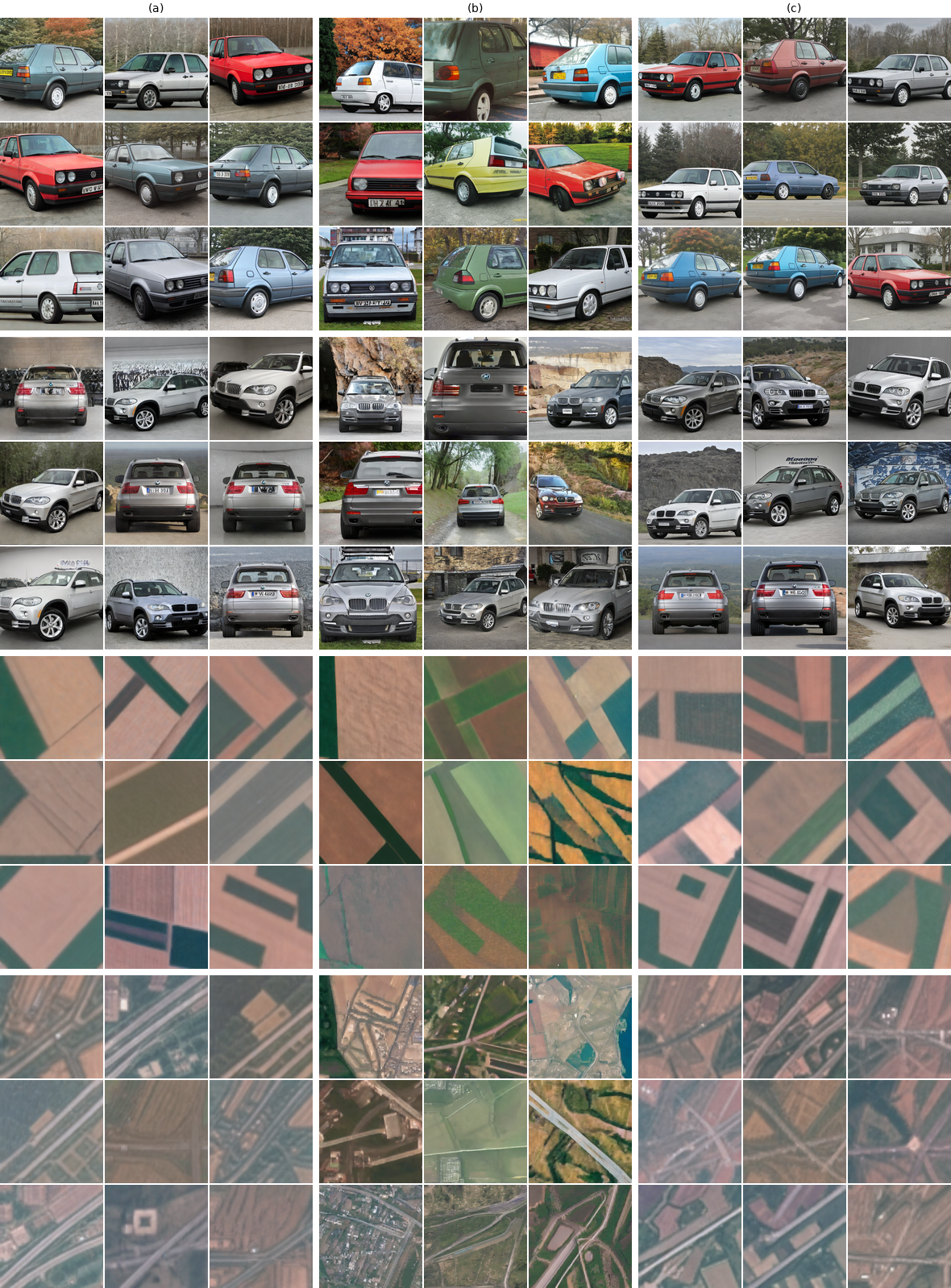}
\caption{
{\em Additional Qualitative Results.} (a) LoFT, (b) DataDream, (c) ChimeraLoRA. From top to bottom, the classes are 1991 Volkswagen Golf Hatchback and 2007 BMW X5 SUV from Cars, followed by annualcrop and highway from EuroSAT.
}
\label{fig:add-1}
\end{figure*}

\begin{figure*}[t!]
\centering
\includegraphics[width=0.8\textwidth]{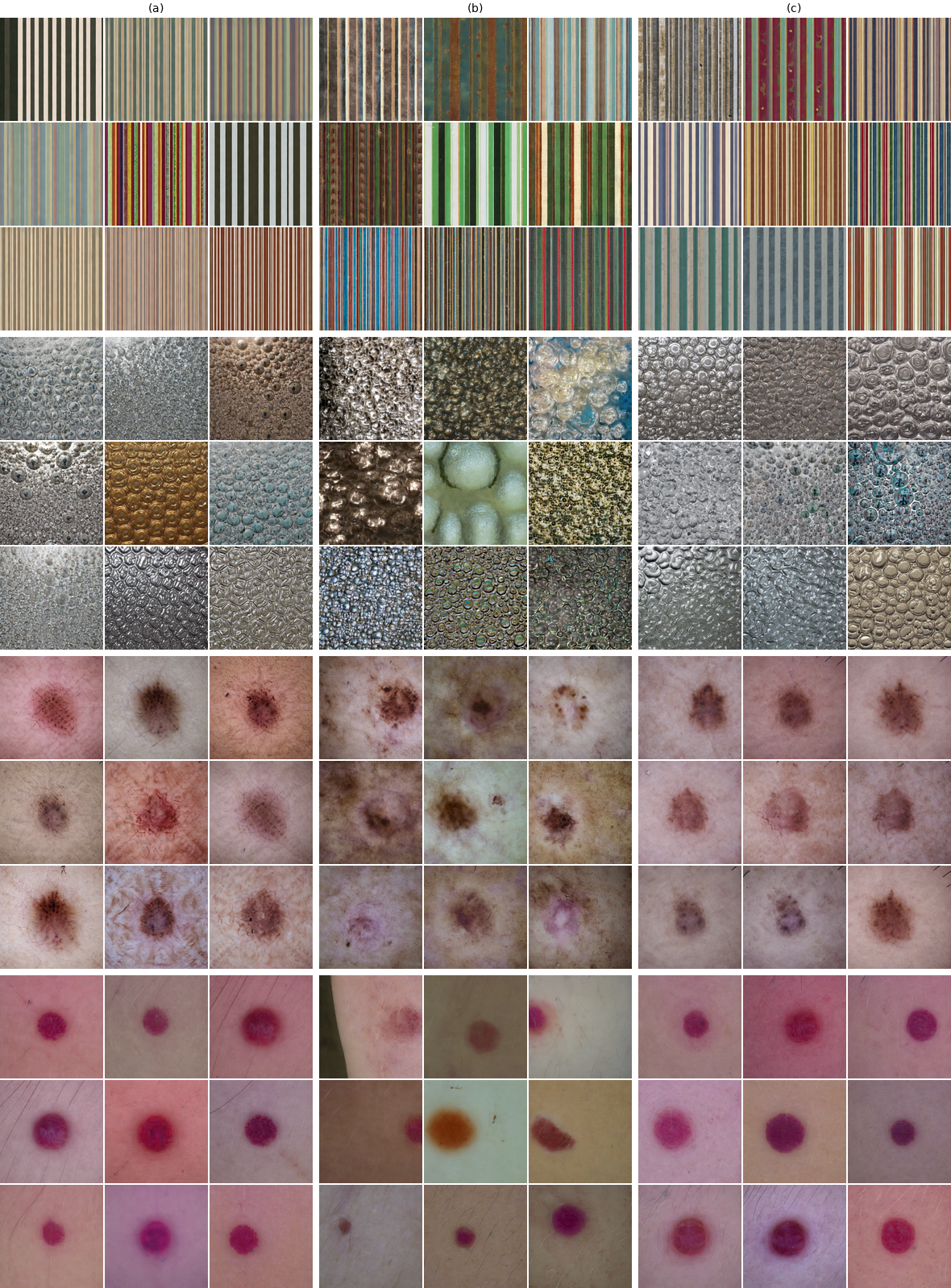}
\caption{
{\em Additional Qualitative Results.} (a) LoFT, (b) DataDream, (c) ChimeraLoRA. From top to bottom, the classes are banded and bubbly from DTD, followed by basal cell carcinoma and vascular lesion from ISIC.
}
\label{fig:add-2}
\end{figure*}

Figures~\ref{fig:add-1} and~\ref{fig:add-2} present additional qualitative results across diverse datasets.

